\pdfoutput=1

\documentclass[11pt]{article}

\usepackage{EACL2023}

\usepackage{graphicx}

\usepackage{times}
\usepackage{latexsym}

\usepackage{lipsum}

\usepackage[T1]{fontenc}

\usepackage[utf8]{inputenc}

\usepackage{microtype}

\usepackage{inconsolata}

\title{Generation, Distillation and Evaluation of Motivational Interviewing-Style Reflections with a Foundational Language Model}

\author{Andrew Brown \ \hspace{0.75cm} Jiading Zhu \ \hspace{0.75cm} Mohamed Abdelwahab \\ {\bf Alec Dong} \ \hspace{0.75cm} {\bf Cindy Wang} \ \hspace{0.75cm} {\bf Jonathan Rose} \\ \\ The Edward S. Rogers Sr. Department of \\ Electrical \& Computer Engineering, University of Toronto \\ \texttt{andrewm.brown@mail.utoronto.ca} \hspace{1cm} \texttt{jonathan.rose@utoronto.ca}}

\begin{document}

\maketitle
\begin{abstract}

Large Foundational Language Models are capable of performing many tasks at a high level but are difficult to deploy in many applications because of their size and proprietary ownership. Many will be motivated to distill specific capabilities of foundational models into smaller models that can be owned and controlled. In the development of a therapeutic chatbot, we wish to distill a capability known as reflective listening, in which a therapist produces \emph{reflections} of client speech. These reflections either restate what a client has said, or connect what was said to a relevant observation, idea or guess that encourages and guides the client to continue contemplation. In this paper, we present a method for distilling the generation of reflections from a Foundational Language Model (GPT-4) into smaller models. We first show that GPT-4, using zero-shot prompting, can generate reflections at near 100\% success rate, superior to all previous methods. Using reflections generated by GPT-4, we fine-tune different sizes of the GPT-2 family. The GPT-2-small model achieves 83\% success on a hold-out test set and the GPT-2 XL achieves 90\% success. We also show that GPT-4 can help in the labor-intensive task of evaluating the quality of the distilled models, using it as a zero-shot classifier. Using triple-human review as a guide, the classifier achieves a Cohen-Kappa of 0.66, a substantial inter-rater reliability figure.

\end{abstract}

\section{Introduction}
Motivational Interviewing (MI) is a counselling technique that is used to guide people towards behaviour change~\cite{miller_motivational_2012}. MI has seen success in smoking cessation~\cite{lindson_motivational_2019} and alcohol consumption reduction~\cite{nyamathi_effect_2010}, among other behaviours. Our long-term goal is to automate MI-based therapeutic conversations in smoking cessation~\cite{brown_motivational_2023}.

\begin{table}
\centering
\begin{tabular}{p{7cm}}
 \hspace{25 mm}\textbf{Conversation} \\
 \hline
 \textbf{MI Clinician:} What are some things you don't like about your smoking addiction?\\
 \textbf{Client:} I don't like making other people uncomfortable with my smoking.\\\\
 \textbf{MI Clinician} (Simple Reflection): You don't enjoy making people feel uncomfortable with your smoking.\\
 \textbf{MI Clinician} (Complex Reflection): You might be feeling self-conscious about your smoking.\\
 \hline
\end{tabular}
\caption{\label{tab:simpleandcomplexreflection}
Example of Simple vs Complex Reflection
}
\end{table}

\begin{figure*}[h]
    \centering
    \includegraphics[scale=0.120]{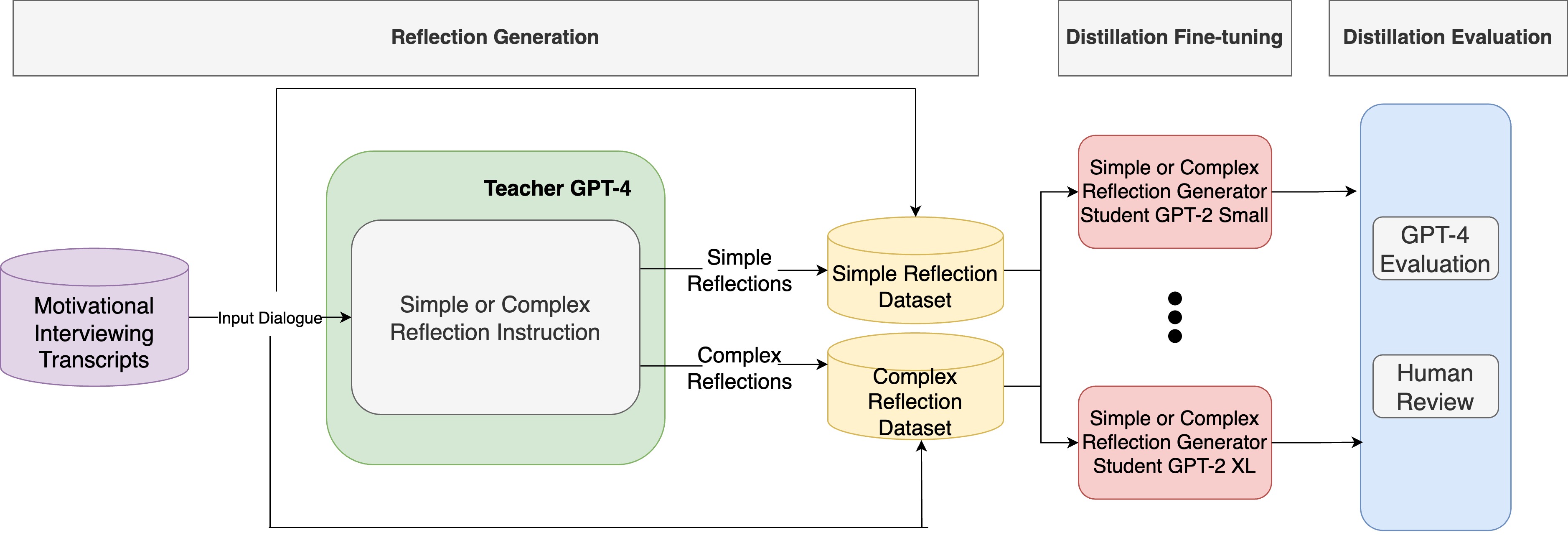}
    \caption{Knowledge Distillation Overview}
    \label{fig:process}
\end{figure*}

A key technique in MI (and many other talk therapies) is \emph{reflective listening}, a conversational approach in which a clinician mirrors the client's thoughts and emotions, enabling them to recognize their own beliefs and contradictions~\cite{miller_motivational_2012}. The core skill of reflective listening is to respond to client utterances with a \emph{reflection}. Reflections are divided into two major types: \emph{simple} reflections which rephrase what a client has said, and \emph{complex} reflections which attempt to infer something based on a recent utterance, or to guess something based on general knowledge~\cite{miller_motivational_2012}. Both types of reflections are illustrated in the conversation snippet in Table~\ref{tab:simpleandcomplexreflection}.

There has been recent work to automate the generation and classification of MI reflections using GPT-2~\cite{radford_language_2019} and GPT-3~\cite{brown_language_2020}.~\cite{ahmed_generation_2022} showed that a few-shot prompted GPT-3 generates MI reflections scoring over 89\% success rate from human annotation and~\cite{shen_counseling-style_2020} demonstrated a fine-tuned GPT-2 generates reflections which are scored by human reviewers as nearly identical to clinician curated reflections. Furthermore,~\cite{ahmed_automatic_2022} showed that a fine-tuned BERT~\cite{devlin_bert_2019} can classify reflections as acceptable at 80\% success rate. In this work we explore the use of zero-shot prompting of GPT-4~\cite{openai_gpt-4_2023} to both generate and classify MI reflections. We use the high-quality reflections from the generator to fine-tune smaller, proprietary models. The latter provides greater privacy for sensitive health communications since the information pathways can be fully controlled when a model is owned by the operator.

In collaboration with MI experts, we designed prompts to generate both simple and complex reflections and classify them with GPT-4. We present a method to distill the reflection generation capability from GPT-4. A dataset was created consisting of questions (that were presented to clients), their answers and generated GPT-4 reflections. These are used to fine-tune smaller language models, and we sought to determine the trade-off between size of the smaller model and its performance. 

In the larger context of a smoking-cessation chatbot that would use the generated reflections, there are situations when a simple reflection is called for, and other times when complex reflection is appropriate~\cite{miller_motivational_2012}. For this reason we will distill two models, one for each type of reflection. 

Figure~\ref{fig:process} illustrates the overall approach used in this work. The fine-tuning datasets are created based on portions of transcripts from a previous chatbot created by the authors~\cite{brown_motivational_2023} and simple or complex reflections generated by GPT-4. Next, as a form of knowledge distillation, we fine-tune the GPT-2~\cite{radford_language_2019} family of models on the simple reflection or complex reflection dataset. To evaluate the student models we employ both human reviewers and use the GPT-4 model itself as a zero-shot classifier. That classification is done in two stages, the first to check for adherency to the principles of MI~\cite{miller_motivational_2012}, and then to classify MI-adherent reflections as simple or complex. The idea of using a large foundational model as an zero-shot evaluator has just begun to appear in the literature~\cite{kamalloo_evaluating_2023, chiang_can_2023} and is not yet well studied. If it can be shown to be successful, it will reduce the costly human effort in determining the effectiveness of distilled and other models. Previous works in MI reflection generation such as~\cite{shen_counseling-style_2020} and~\cite{ahmed_automatic_2022} have used human curated datasets to train classifiers.

The contributions of this paper are: (1) State-of-the-art success rate in generation of reflections; (2) an example of end-to-end task-specific distillation from a foundational language model; and (3) demonstration of the effectiveness of using a foundational language models to evaluate reflections, which has the potential to reduce the amount of human labour in generative model work.

\section{Related Work}
\textbf{Generative Reflections}\\ There have been past attempts to generate MI reflections using transformer-based language models. The work in~\cite{shen_counseling-style_2020} showed that GPT-2 could generate counseling-style MI reflections by fine-tuning on the dialogue context and responses retrieved from similar counseling sessions. Human reviewers scored a test set of generated reflections at 4.13 on a 5-point likert scale while scoring known-good reflections at 3.84, suggesting that the human reviewers preferred the quality of generated reflections over known-good ones. These reflections were proposed to be used in clinician training, allowing for easier access to context specific reflections. This work was subsequently improved by including commonsense and domain specific knowledge while generating responses, similar to what counselors do~\cite{shen_knowledge_2022}. These generated reflections scored lower on human review scores. On reflection coherence, accuracy and preference, human reviewers scored ground-truth reflections higher than generated domain specific reflections.

~\cite{ahmed_generation_2022} investigated the use of prompting and fine-tuning transformer-based language models to generate and classify MI reflections for smoking cessation. Human reviewers scored reflection acceptability on a prompted GPT-2 XL as 54\%, a prompted GPT-3 as 89\%, and a fine-tuned GPT-2 XL at 80\%. For reflection classification,~\cite{ahmed_automatic_2022} fine-tuned a BERT model to achieve 81\% accuracy in classifying reflections.

We view the previous work in MI reflection generation and classification as preliminary and seek to build upon it. With GPT-4, our goal is to create an improved reflection generation which scores higher with human reviewers than that of~\cite{shen_knowledge_2022} and~\cite{ahmed_automatic_2022}, and create a more accurate reflection classifier than~\cite{ahmed_automatic_2022} which agrees with human decisions. 
\\
\textbf{Knowledge Distillation}\\
Knowledge distillation is a technique in machine learning where a student model is trained to
reproduce the behaviour of a teacher model, typically to achieve model compression~\cite{gu_knowledge_2023}.~\cite{hinton_distilling_2015} showed the first method of knowledge distillation in which a student neural network was trained to mimic a teacher model's performance on MNIST and speech recognition. The student was trained using a loss function which optimized a combined objective of minimizing the loss of the ground-truth labels and the teacher model's output logits as labels.

Knowledge distillation has since been successfully applied to language models, with DistilBERT~\cite{sanh_distilbert_2020}, a transformer-based language model trained using a loss function for the student model similar to~\cite{hinton_distilling_2015} for the purpose of compressing BERT~\cite{devlin_bert_2019}. Subsequently, researchers have also considered Task-specific knowledge distillation, which seeks to distill a subset of the teacher model's capability into the student. Two examples of this are ~\cite{tang_distilling_2019} which sought to distill only sentiment analysis, and~\cite{liu_rethinking_2022} which focused on the tasks specific to the GLUE dataset benchmark~\cite{wang_glue_2019}.

Other knowledge distillation works use different loss functions during training, while others employed pre-trained models as the student.~\cite{he_generate_2022} showed a method for task-specific knowledge distillation using pre-trained transformer language models as the student and fine-tuning for training. First, a teacher language model is instructed to generate a dataset of additional prompts and output text using an initial set of prompts. Next, this dataset is annotated for data quality and used to fine-tune the smaller student models.

The Self-Instruct approach~\cite{wang_self-instruct_2023} is another application of knowledge distillation which fine-tuned a pre-trained language model. First, a set of 175 seed prompts (describing text instructions for many tasks) were created and used to generate more instructions using GPT-3~\cite{brown_language_2020}. Next, GPT-3 also generates inputs for the instructions and then the corresponding output. This creates a text dataset of instructions, inputs and outputs. Finally, the dataset is used to fine-tune GPT-3, the same model which generated the dataset. Motivated by Self-Instruct,~\cite{alpaca} created Alpaca, an instruction following LLaMA~\cite{touvron_llama_2023} language model created through fine-tuning on text generated by InstructGPT. The Alpaca method also uses GPT-3 to generate a knowledge distillation dataset, but shrinks the student architecture to the LLaMA-7B model~\cite{touvron_llama_2023}, a compression of 25 times. Alpaca's quality of generation were shown to be close to the GPT-3 teacher model, showing that this method of knowledge distillation through generated text can be used to create models a fraction of the size with competitive performance.

The present work combines ideas from previous research in generative MI reflections and knowledge distillation. We use a style of zero-shot prompting similar to~\cite{wang_self-instruct_2023} with GPT-4 to generate MI reflections with the same goal as~\cite{shen_counseling-style_2020} and~\cite{ahmed_automatic_2022}. Next, we distill knowledge by fine-tuning smaller transformer-based language models similar to~\cite{he_generate_2022}. It is important to acknowledge that our method of knowledge distillation is different from the recent works in~\cite{hinton_distilling_2015,sanh_distilbert_2020, devlin_bert_2019}. We use the term distillation as it most accurately describes the underlying task of transferring knowledge from a large model to a smaller one.
\section{Method}
The goals of this paper are to generate high-success rate reflections using GPT-4, to distill that capability into smaller models and measure their success rate, and to determine how well a zero-shot prompt-based GPT-4 model can evaluate the quality of reflections. This section describes the methods for each of these steps.

\subsection{Dataset Collection}
\label{sec:gatheringmotivationalinterviewingtranscripts}
To generate MI reflections from GPT-4, we need input questions and answers from a MI conversation. Mentioned previously, we use transcripts from the smoking cessation MI chatbot created by the authors~\cite{brown_motivational_2023}. Table~\ref{tab:convo} shows an excerpt of a conversation transcript. The chatbot adopts a pattern of asking open-ended questions (QUESTION), retrieving answers (ANSWER), and generating reflections (REFLECTION) as shown in Table~\ref{tab:convo}. We gather question and answers without the reflection as inputs to generate a reflection with GPT-4. In total, 4194 question-answer pairs are divided into 2394 training set examples, 599 validation set examples, and 1201 holdout testing set examples. Each question-answer pair has a simple and complex reflection generated, thus totalling 8388 dataset entries (but models are only trained, validated, and tested on the 4194 dataset entries with simple reflections or 4194 dataset entries with complex reflections).

\subsection{Reflection Generation with GPT-4}
\label{sec:refgen}
Reflection generation is done using zero-shot prompting with GPT-4. We use the question-answer pairs described in Section~\ref{sec:gatheringmotivationalinterviewingtranscripts} with a prepended instruction to generate either a simple or complex reflection. The input prompt and reflection are gathered into a dataset, and used to fine-tune student models, as discussed in Section~\ref{sec:finetuningknowledgedistillatonprocess}.

The instruction for simple and complex reflection generation prompts were developed iteratively on a private test set. First, we hand-wrote an initial prompt and tested it on just a few (1-5) examples. We then increased the size of the test set, noting the examples in which the prompt generated non-MI-adherent reflections, and made modifications accordingly. While evolving the prompt we prioritized maintaining its generality, ensuring that the language use would accommodate many examples, rather than just a few specific ones. For example, in one of the iterations, we noticed that a few generated reflections included questions rather than statements, making these reflections non-MI-adherent. The prompt was modified by adding the sentence "The reflection must be a statement and not a question", which is a general instruction. Throughout this iterative design process, we also consulted with MI experts to get feedback and suggestions on the wording of the prompt.

\begin{figure*}[h]
    \centering
    \includegraphics[scale=0.15]{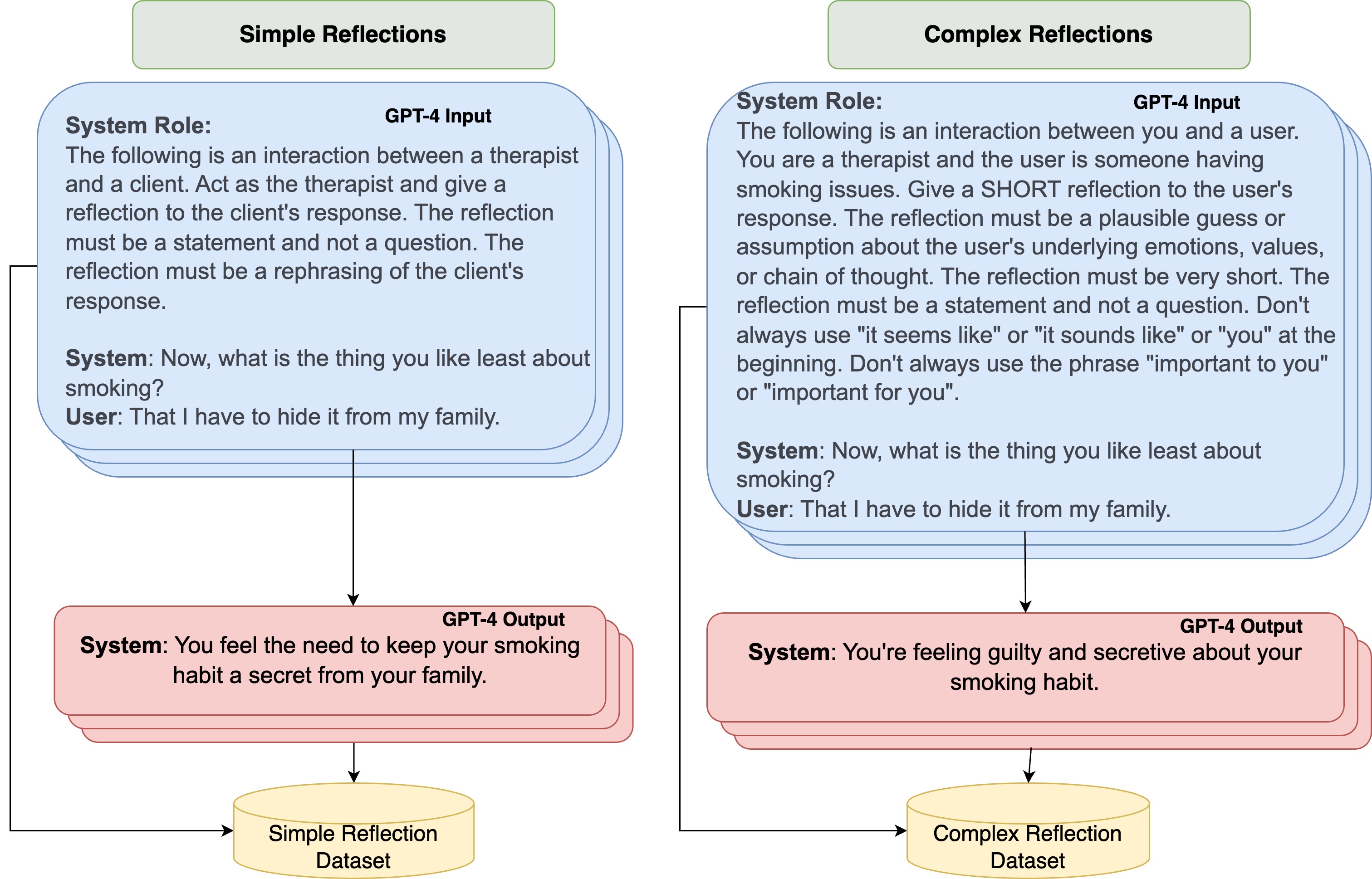}
    \caption{Reflection Data Generation}
    \label{fig:refgen}
\end{figure*}

The full prompt for generating reflections with GPT-4 uses OpenAI's chat-complete~\cite{openai_gpt-4_2023} format, which divides the input prompt into three segments:~\textit{System Role},~\textit{System Message}, and~\textit{User Message}. The~\textit{System Role} is the instruction of the desired task, which in this work is the prompt for generating a simple or complex reflection. The~\textit{System Message} and~\textit{User Message} are questions and answers, respectively, from our MI dataset like the one seen in Table~\ref{tab:convo}. Figure~\ref{fig:refgen} shows the full prompt for simple and complex reflection generation, with an example for each. Additionally, the prompt for simple and complex reflection generation can be viewed by itself in Appendix~\ref{app:gpt-4prompts}. Hereinafter, we refer to a prompted GPT-4 for reflections as the GPT-4 Reflection Generator.

We perform a separate validation of the GPT-4 Reflection Generator through a human review. This is described in Section~\ref{sec:humanreview}. 

\begin{table}
\centering
\begin{tabular}{p{7cm}}
 \hspace{30 mm}\textbf{Context} \\
 \hline
 \textbf{Bot:} (QUESTION) To start, what is the thing you like most about smoking?\\
 \textbf{Client:} (ANSWER) Stress relief.\\
 \textbf{Bot:} (REFLECTION) You enjoy smoking because it helps you cope with stressful situations.\\
 \textbf{Bot:} Did that make sense?\\
 \textbf{Client:} Yes.\\
 \textbf{Bot:} That's great to hear, thanks for letting me know!\\
 \textbf{Bot:} (QUESTION) Now, what is the thing you like least about smoking?\\
 \textbf{Client:} (ANSWER) I spend a lot of money on cigarettes.\\
 \textbf{Bot:} (REFLECTION) You dislike spending money on cigarettes.\\
 \hspace{25 mm}· · · (more turns)\\
    
 \hline
\end{tabular}
\caption{\label{tab:convo}
MI Chatbot Transcript Excerpt 
}
\end{table}

\begin{figure*}[h]
    \centering
    \includegraphics[scale=0.10]{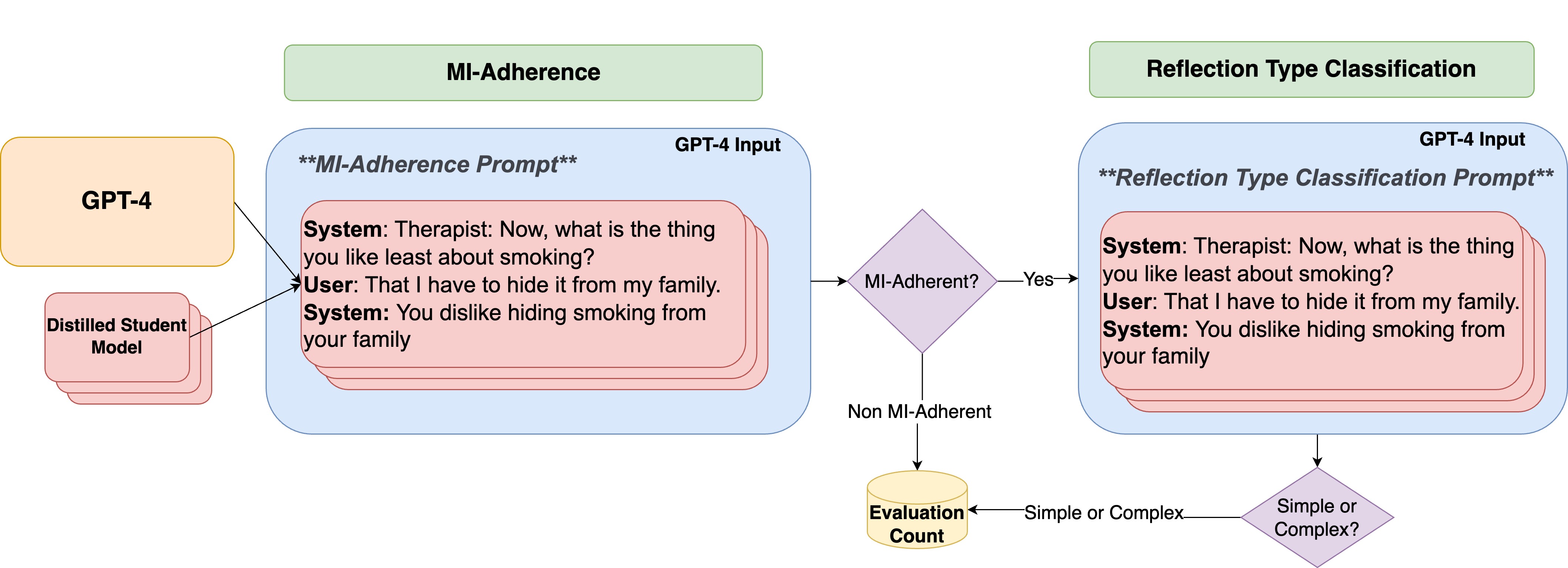}
    \caption{Reflection Evaluation Pipeline}
    \label{fig:refevaluate}
\end{figure*}

\subsection{Fine-tuning Knowledge Distillation Process}
\label{sec:finetuningknowledgedistillatonprocess}
After gathering the dataset of MI conversation questions, answers, and GPT-4-generated reflections, we use fine-tuning to distill that reflection capability in a student model. We motivate this method by noting that state-of-the-art foundational language models such as GPT-4~\cite{openai_gpt-4_2023} do not provide access to the output logits or probabilities used in next word prediction, which are required in a distillation method such as~\cite{hinton_distilling_2015}. Furthermore, it has been shown in recent research~\cite{hwang_comparison_2022} that using specific labels rather the soft logit target for distillation can be more effective when the student-teacher architectures are very different, which is likely true between GPT-4 and GPT-2. Below we describe the text formatting used and details of fine-tuning.

Table~\ref{tab:triplet} in Appendix~\ref{app:appendixfinetuningtextformat} shows example fine-tuning entries for simple and complex reflections. The text that the student model is trained on consists of the appropriate prompt (described above, either simple or complex) followed by the question, answer, and reflection. We use a triple \# sign to separate the instruction and conversation, as suggested in the fine-tuning data for the Alpaca language model~\cite{alpaca}.

\subsection{Student Model Selection}
\label{sec:studentmodels}
We selected the GPT-2~\cite{radford_language_2019} family transformer-based language models as students. The GPT-2 family was selected because of the open source status of the models, range in architecture size, and demonstration in past works for reflection generation. All models have been pre-trained on the WebText dataset, a 40GB corpus of diverse text. We investigated how the different model sizes in the GPT-2 family affects the knowledge distillation outcome. The GPT-2 family has a large variety of sizes, with the smallest to the largest being an increase of 12 times. 



\subsection{Reflection Evaluation with GPT-4}
\label{sec:gpt4eval}
To evaluate reflections, we use a zero-shot prompt-based GPT-4~\cite{openai_gpt-4_2023} in two ways:
\begin{enumerate}
    \item MI-Adherence: Classify the reflection as MI-adherent~\cite{miller_motivational_2012} or not. Reflections classified as not MI-adherent are not sent to step two. This classifier checks if the reflection abides by the principles of MI. This is the most basic qualification of an MI reflection and gives an indication of how well the reflection model is performing.
    \item Reflection Type Classification: Classify the reflection as Simple or Complex. We know that it is possible for the simple generator to produce complex classifications and vice-versa.
\end{enumerate}

Figure~\ref{fig:refevaluate} illustrates the evaluation pipeline explained above. Furthermore, the MI-adherence prompt and reflection type classification prompt can be be seen in Table~\ref{tab:gpt-4prompts} in Appendix~\ref{app:gpt-4prompts}. 

As described in Section~\ref{sec:refgen}, the design of each prompt for evaluation was done through human-based evolution and testing using a private test-set in collaboration with MI experts. Each prompt was hand-written and evolved until we were able to reach an acceptable success rate on a test-set, then the size of the test-set was increased. This process repeated until we were satisfied with the overall performance.

We provide a separate measurement of the performance of each prompt in Section~\ref{sec:agreementbetweengpt-4andhumanreview} by recruiting human annotators to also classify MI-adherence and reflection type classification, then calculating the Cohen kappa~\cite{mchugh_interrater_2012} on the classifications. The Cohen kappa~\cite{mchugh_interrater_2012, cohen_coefficient_1960}, is a validated metric to measure inter-rater reliability between multiple reviewers (in this case GPT-4 and humans). The score ranges from -1 to 1 representing perfect disagreement and agreement and any score of 0.6 or above is considered substantial~\cite{mchugh_interrater_2012}.

\begin{table*}[htbp]
\centering
\begin{tabular}{lp{1cm}|p{1.1cm}p{0.85cm}|p{1.2cm}p{1.25cm}|p{1.25cm}p{0.8cm}}
\hline\hline
  &  & \multicolumn{2}{c}{MI-adherence} & \multicolumn{2}{c}{Classified Simple} & \multicolumn{2}{c}{Classified Complex} \\
Model - Task          & Size & GPT-4 & HR & GPT-4 & HR & GPT-4 & HR\\
\hline\hline
GPT-2 Small - Simple  & 124M       & 0.76                       & 0.90                          & 0.78                    & 0.69                    & 0.22                      & 0.31                        \\
GPT-2 Medium - Simple & 355M       & 0.91                       & 0.87                          & 0.77                     & 0.81                    & 0.23                      & 0.19                         \\
GPT-2 Large - Simple  & 774M       & 0.93                       & 0.90                          & 0.79                  & 0.71                     & 0.21                      & 0.29                         \\
GPT-2 XL - Simple     & 1.5B       & \textbf{0.93}                       & \textbf{0.92}                          & \textbf{0.80}                     & \textbf{0.82}                       & 0.20                      & 0.18                         \\
\hline
\color{blue} GPT-4 - Simple        & $>>>$    & \textbf{0.99}                       & \textbf{1.00}                          & \textbf{0.91}                    & \textbf{0.97}                         & 0.08                      & 0.03                          \\
\hline
GPT-2 Small - Complex  & 124M       & 0.83                       & 0.85                          & 0.25                     & 0.17                      & 0.76                      & 0.83                         \\
GPT-2 Medium - Complex & 355M       & 0.86                       & 0.92                          & 0.25                     & 0.05                      & 0.75                      & \textbf{0.95}                         \\
GPT-2 Large - Complex  & 774M       & 0.86                       & \textbf{0.97}                          & 0.23                   & 0.17                      & \textbf{0.77}                      & 0.83                         \\
GPT-2 XL - Complex     & 1.5B       & \textbf{0.90}                       & 0.92                          & 0.26                      & 0.11                        & 0.74                      & 0.89                         \\
\hline
\color{blue} GPT-4 - Complex        & $>>>$   & \textbf{0.98}                       & \textbf{1.00}                          & 0.26                      & 0.13                        & \textbf{0.74}                      & \textbf{0.87}   \\
\hline\hline
\end{tabular}
\caption{\label{tab:MI-adherence-Task-cls-results}
MI-adherence and reflection type classification scores of distilled student models and teacher GPT-4 Reflection Generator. HR stands for Human Review.
}
\end{table*}

\subsection{Human Review}
\label{sec:humanreview}
We recruited five annotators to evaluate reflections from GPT-4 and each distilled student model. The five annotators consist of four males and one female at an average age of 23, located in North America. Each annotator has a basic understanding of MI having read~\cite{miller_motivational_2012} and taken coursework~\footnote{http://test.teachdev.ca/ola/index.html}.~\cite{wu_are_2023} observed that lay-people are able to label MI reflections with consistent inter-group correlation.

From the holdout-set of 1201 examples with reflections, 61 ($\sim$5\%) are randomly sampled with stratification\footnote{Reflections were stratified by the question asked, to ensure there is diverse context.} from each model for human review. We review 10 models in total: the GPT-4 Reflection Generator for simple and complex reflections and four student GPT-2 models of different sizes for simple and complex reflections. This gives a total of 610 review examples. 

The human review process closely follows the same two step pipeline for reflection review as explained in Section~\ref{sec:gpt4eval}: For MI-adherence, annotators classify reflections using their own understanding of MI. For reflection type classification, annotators classify reflections as either simple or complex. Reflections are assumed as simple unless there is a plausible assumption about the client's underlying emotions, values, or chain of thought, similar to the prompt created for complex reflections in Figure~\ref{fig:refgen}.

Three annotators independently make a binary decision for MI-adherence, and the majority from the three choices is taken. Next, if the reflection is MI-adherent, then the three annotators make another binary decision of reflection type classification and the majority result, from the three, is chosen. We use the two aggregate decisions to calculate the agreement score explained in Section~\ref{sec:gpt4eval}.

\section{Experiment}
\subsection{Experimental Setup}
\label{sec:experimentsetup}
The four GPT-2 student models are implemented using PyTorch~\cite{paszke_pytorch_2019} and were acquired from the HuggingFace Transformers library~\cite{wolf_huggingfaces_2020}. Training and inference was done using 4 NVIDIA A10G Tensor Core GPUs and used DeepSpeed ZeRO~\cite{rajbhandari_zero_2020} parallelism and CPU offloading. All models were trained using a hyperparameter search. We searched for Batch Size in [8, 16, 32, 64] and Learning Rate in [0.00005, 0.0005, 0.001]. The chosen hyperparameters are given in Appendix~\ref{app:hyperparameters}, Table~\ref{tab:hyperparameters}. All fine-tuning used 4 epochs with early stopping. We used the Adam Optimizer~\cite{kingma_adam_2017} with zero weight decay. For inference, we used decoding parameters as temperature=0.6 with top-k=100 and top-p=1.0. The code used to train and test models can be found here.\footnote{https://github.com/andrewmbrown/transformer-fine-tune}

\section{Results and Analysis}
\label{sec:ch5intro}
In this section we report the quality (using human review) of the reflections generated by the GPT-4 Reflection Generator. Then, we compare the quality of the automatic evaluation using GPT-4 (with the evaluation prompt, described in Section~\ref{sec:gpt4eval}) with human review. Finally, we present and discuss the performance of distilled GPT-2 models. 

The generation and evaluation results for all of these models are given in one large table, Table~\ref{tab:MI-adherence-Task-cls-results}, but are discussed separately in Section~\ref{sec:performanceofgpt-4reflections} and Section~\ref{sec:mi-adherenceandscopeclassification}. Each of the values in Table~\ref{tab:MI-adherence-Task-cls-results} gives the fraction of the test set that was deemed acceptable by the evaluation method. For example, the 0.99 score in MI-Adherence for the GPT-4 simple Reflection Generator indicates that 99\% of the 1201 generated simple reflections were judged as MI-adherent by the GPT-4 MI-Adherence classifier. The right-most four columns of Table~\ref{tab:MI-adherence-Task-cls-results} give the fraction of the reflections that were deemed, by the GPT-4 Reflection Type Classifier or the human review, to be a simple reflection or complex reflection. Student models are listed in each row of the table, in order of increasing model size, and are grouped by which reflection generation task they performed - simple or complex. The table also includes the results from the GPT-4 Reflection Generator in blue. To find the number of examples used to calculate reflection type classification scores, multiply the original set size (1201 for GPT-4 and 61 for human review) by the respective MI-adherence score (reflections must first be MI-adherent before reflection type classification as mentioned in Section~\ref{sec:gpt4eval}). Additionally, the precision, recall, and F1 scores for evaluation done by GPT-4 is given in Appendix~\ref{app:precisionrecallf1}.

\subsection{GPT-4 Reflection Generation}
\label{sec:performanceofgpt-4reflections}
Rows 6 and 11 (with blue text) of Table~\ref{tab:MI-adherence-Task-cls-results} give the scores of the prompted (simple and complex) GPT-4 Reflection Generator, and we focus here only on the human review (HR) columns. A key result is that the GPT-4 Reflection Generator achieves a 100\% success rate on MI-adherence, for both simple and complex reflections. This is much better than prior work on reflection generation, which achieved 89\% using GPT-3~\cite{ahmed_automatic_2022} and 4.13/5 in~\cite{shen_counseling-style_2020}.
This success makes it a candidate for distillation, and indeed is what motivated the present work. 

The simple prompted GPT-4 reflections were labelled as simple 97\% of the time, while the complex reflections were deemed as complex 87\% of the time. For those that were not complex, it may have been because the client response itself was not amenable to a complex reflection.

\begin{table}[htbp]
\centering
\begin{tabular}{lll}
\hline
Task   & MI-A & RT-CLS \\ \hline
Simple & 0.671                           & 0.604                                                          \\
Complex & 0.429                           & 0.711                                                       \\
\hline
All & 0.54                            & 0.66                                                            \\ \hline
\end{tabular}
\caption{\label{tab:cohenkappa}
Inter-Rater Reliability Cohen kappa scores between GPT-4 and Human Reviewers on three evaluation tasks. MI-A and RT-CLS refer to Motivational Interviewing Adherence and Reflection Type Classification respectively.
}
\end{table}

\subsection{GPT-4 Reflection Classification}
\label{sec:agreementbetweengpt-4andhumanreview}
Section~\ref{sec:gpt4eval} describes a method for using GPT-4 to evaluate the quality of reflections produced by models, as an alternative to laborious human review. In this section we compare it to human review, using the Cohen kappa Inter-Rater Reliability~\cite{mchugh_interrater_2012, cohen_coefficient_1960} coefficient. Table~\ref{tab:cohenkappa} presents the Cohen kappa coefficient between GPT-4-based evaluation and human evaluation for MI-adherence (MI-A) and reflection type classification (RT-CLS). Within each column, the agreement is shown individually for simple and complex reflections, with a final value combining both types of reflections in the last row.

The Cohen kappa scores are calculated on the samples that overlap between the larger 1201 entry holdout set used to test the GPT-4 based method, and the 61 entry holdout set used in human review. For MI-adherence, the simple and complex reflection kappa is calculated on 305 examples each (61 examples for five models) and the final row is calculated on 610. Reflection type classification scores are calculated on 272 examples for simple reflections and 261 for complex reflections giving a total of 533 in the combined row.

Table~\ref{tab:cohenkappa} shows that there is substantial agreement (0.671) between human and GPT-4 based classification of the generated simple reflections classification for MI-adherence. There is near substantial agreement for complex reflection classification (0.429). Overall, the bottom row kappa of 0.54 suggests that there is near substantial agreement between the GPT-4 classifier and human review, validating our use of GPT-4 for MI-adherence.

For reflection type classification, we observe substantial agreement for simple reflections, complex reflections, and the combined final row. This validates our use of GPT-4 for reflection type classification as we observe substantial agreement on all tasks. 
\subsection{Performance of Distilled Reflection Generation Models}
\label{sec:mi-adherenceandscopeclassification}
In this section we discuss the results of student models shown in Table~\ref{tab:MI-adherence-Task-cls-results}.
\\
\textbf{MI-Adherence}: The third and fourth column of Table~\ref{tab:MI-adherence-Task-cls-results} show MI-adherence scores. In almost every case the result is superior to the success rate achieved by~\cite{ahmed_automatic_2022} for a fine-tuned GPT-2-XL model (which achieve an 80\% success rate). Our method creates both a simple and complex GPT-2 Medium reflector which scores higher in MI-adherence while being four times smaller that the GPT-2 XL of~\cite{ahmed_automatic_2022}. Furthermore, as model size increases, MI-adherence scores increase.
\\
\textbf{Reflection Type Classification}: The 5th, 6th, 7th, and 8th columns of Table~\ref{tab:MI-adherence-Task-cls-results} give reflection type classification scores for distilled simple and complex reflection models. The distilled simple reflection generation models are almost as good as the simple GPT-4 Reflection Generator are at producing simple reflections. The distilled complex reflection generation models are as good as the complex GPT-4 Reflection Generator at producing complex reflections.

\section{Conclusion}
We have presented a method for generating simple and complex MI reflections using GPT-4, and shown that it is capable of near-perfect success, beyond the previous state of the art. We showed how to distill those capabilities into to smaller, GPT-2-based student models, and that the range of sizes results in success rates ranging from 76\% to 93\%. One issue in distillation work is the labour to determine the success of the distilled models; we have shown that a classification prompt with GPT-4 as an evaluator is reliable. This paper provides a case study of distillation of a specific task from an expensive, privacy-challenged large foundational model into an owned, smaller pre-trained language model.

\section*{Limitations}
The results presented are specific to the example dataset that we have used, and may not generalize to other kinds of reflections, as mentioned in Section~\ref{sec:gatheringmotivationalinterviewingtranscripts}. Also, the evaluation techniques described in Section~\ref{sec:experimentsetup}, used a much smaller size of holdout set for the human review (compared to the holdout set using the GPT-4-based review). This was done in order to reduce the labour of labelling, but results in a smaller sub-set which is less accurate.

Finally, the reflection classification process for human reviewers presented in Section~\ref{sec:humanreview} may not accurately capture what it means to generate an acceptable reflection. Previously mentioned works like~\cite{shen_counseling-style_2020} and~\cite{shen_knowledge_2022} used specific qualities of a reflection like coherence, accuracy, and preference, while our work mainly uses MI-adherence. In future work we aim to incorporate these criteria for a more complex reflection classification.

\section*{Ethics Statement}
We guarantee that the data we gather for reflection generation comes from experiments that users have willingly participated in, and the overall process received ethics board approval. All human reviewers were recruited through local word-of-mouth contact and were fairly compensated for their time. Collected and generated data was reviewed to ensure personally identifiable or sensitive information was removed.

We also guarantee that all our deployment of generative language models for reflection generation is approved under an ethics board. Using generative language models for reflection generation in a chatbot has associated risks. Inaccurate or inappropriate reflections are capable of moving individuals with addictions even farther away from healthy behaviour change~\cite{miller_motivational_2012}. 
\bibliography{anthology,custom}

\begin{thebibliography}{32}
\expandafter\ifx\csname natexlab\endcsname\relax\def\natexlab#1{#1}\fi

\bibitem[{Ahmed(2022)}]{ahmed_automatic_2022}
Imtihan Ahmed. 2022.
\newblock \href {https://tspace.library.utoronto.ca/handle/1807/123170} {\emph{Automatic {Generation} and {Detection} of {Motivational} {Interviewing}-style {Reflections} for {Smoking} {Cessation} {Therapeutic} {Conversations} using {Transformer}-based {Language} {Models}}}.
\newblock Thesis.
\newblock Accepted: 2022-06-29T15:11:56Z.

\bibitem[{Ahmed et~al.(2022)Ahmed, Rose, Keilty, Cooper, and Selby}]{ahmed_generation_2022}
Imtihan Ahmed, Jonathan Rose, Eric Keilty, Carolynne Cooper, and Peter Selby. 2022.
\newblock \href {https://doi.org/10.36227/techrxiv.20029880.v1} {Generation and {Classification} of {Motivational}-{Interviewing}-{Style} {Reflections} for {Smoking} {Behaviour} {Change} {Using} {Few}-{Shot} {Learning} with {Transformers}}.

\bibitem[{Brown et~al.(2023)Brown, Kumar, Melamed, Ahmed, Wang, Deza, Morcos, Zhu, Maslej, Minian, Sujaya, Wolff, Doggett, Iantorno, Ratto, Selby, and Rose}]{brown_motivational_2023}
Andrew Brown, Ash~Tanuj Kumar, Osnat Melamed, Imtihan Ahmed, Yu~Hao Wang, Arnaud Deza, Marc Morcos, Leon Zhu, Marta Maslej, Nadia Minian, Vidya Sujaya, Jodi Wolff, Olivia Doggett, Mathew Iantorno, Matt Ratto, Peter Selby, and Jonathan Rose. 2023.
\newblock \href {https://doi.org/10.2196/49132} {A {Motivational} {Interviewing} {Chatbot} {With} {Generative} {Reflections} for {Increasing} {Readiness} to {Quit} {Smoking}: {Iterative} {Development} {Study}}.
\newblock \emph{JMIR Ment Health}, 10:e49132.

\bibitem[{Brown et~al.(2020)Brown, Mann, Ryder, Subbiah, Kaplan, Dhariwal, Neelakantan, Shyam, Sastry, Askell, Agarwal, Herbert-Voss, Krueger, Henighan, Child, Ramesh, Ziegler, Wu, Winter, Hesse, Chen, Sigler, Litwin, Gray, Chess, Clark, Berner, McCandlish, Radford, Sutskever, and Amodei}]{brown_language_2020}
Tom~B. Brown, Benjamin Mann, Nick Ryder, Melanie Subbiah, Jared Kaplan, Prafulla Dhariwal, Arvind Neelakantan, Pranav Shyam, Girish Sastry, Amanda Askell, Sandhini Agarwal, Ariel Herbert-Voss, Gretchen Krueger, Tom Henighan, Rewon Child, Aditya Ramesh, Daniel~M. Ziegler, Jeffrey Wu, Clemens Winter, Christopher Hesse, Mark Chen, Eric Sigler, Mateusz Litwin, Scott Gray, Benjamin Chess, Jack Clark, Christopher Berner, Sam McCandlish, Alec Radford, Ilya Sutskever, and Dario Amodei. 2020.
\newblock \href {https://doi.org/10.48550/arXiv.2005.14165} {Language {Models} are {Few}-{Shot} {Learners}}.
\newblock ArXiv:2005.14165 [cs].

\bibitem[{Chiang and Lee(2023)}]{chiang_can_2023}
Cheng-Han Chiang and Hung-yi Lee. 2023.
\newblock \href {https://doi.org/10.18653/v1/2023.acl-long.870} {Can {Large} {Language} {Models} {Be} an {Alternative} to {Human} {Evaluations}?}
\newblock In \emph{Proceedings of the 61st {Annual} {Meeting} of the {Association} for {Computational} {Linguistics} ({Volume} 1: {Long} {Papers})}, pages 15607--15631, Toronto, Canada. Association for Computational Linguistics.

\bibitem[{Cohen(1960)}]{cohen_coefficient_1960}
Jacob Cohen. 1960.
\newblock \href {https://doi.org/10.1177/001316446002000104} {A {Coefficient} of {Agreement} for {Nominal} {Scales}}.
\newblock \emph{Educational and Psychological Measurement}, 20(1):37--46.
\newblock Publisher: SAGE Publications Inc.

\bibitem[{Devlin et~al.(2019)Devlin, Chang, Lee, and Toutanova}]{devlin_bert_2019}
Jacob Devlin, Ming-Wei Chang, Kenton Lee, and Kristina Toutanova. 2019.
\newblock \href {https://doi.org/10.48550/arXiv.1810.04805} {{BERT}: {Pre}-training of {Deep} {Bidirectional} {Transformers} for {Language} {Understanding}}.
\newblock ArXiv:1810.04805 [cs].

\bibitem[{Gu et~al.(2023)Gu, Dong, Wei, and Huang}]{gu_knowledge_2023}
Yuxian Gu, Li~Dong, Furu Wei, and Minlie Huang. 2023.
\newblock \href {http://arxiv.org/abs/2306.08543} {Knowledge {Distillation} of {Large} {Language} {Models}}.
\newblock ArXiv:2306.08543 [cs].

\bibitem[{He et~al.(2022)He, Nassar, Kiros, Haffari, and Norouzi}]{he_generate_2022}
Xuanli He, Islam Nassar, Jamie Kiros, Gholamreza Haffari, and Mohammad Norouzi. 2022.
\newblock \href {https://doi.org/10.48550/arXiv.2106.06168} {Generate, {Annotate}, and {Learn}: {NLP} with {Synthetic} {Text}}.
\newblock ArXiv:2106.06168 [cs].

\bibitem[{Hinton et~al.(2015)Hinton, Vinyals, and Dean}]{hinton_distilling_2015}
Geoffrey Hinton, Oriol Vinyals, and Jeff Dean. 2015.
\newblock \href {http://arxiv.org/abs/1503.02531} {Distilling the {Knowledge} in a {Neural} {Network}}.
\newblock ArXiv:1503.02531 [cs, stat].

\bibitem[{Hwang et~al.(2022)Hwang, Sim, Zhang, and Strohman}]{hwang_comparison_2022}
Dongseong Hwang, Khe~Chai Sim, Yu~Zhang, and Trevor Strohman. 2022.
\newblock \href {http://arxiv.org/abs/2210.05793} {Comparison of {Soft} and {Hard} {Target} {RNN}-{T} {Distillation} for {Large}-scale {ASR}}.
\newblock ArXiv:2210.05793 [cs, eess].

\bibitem[{Kamalloo et~al.(2023)Kamalloo, Dziri, Clarke, and Rafiei}]{kamalloo_evaluating_2023}
Ehsan Kamalloo, Nouha Dziri, Charles Clarke, and Davood Rafiei. 2023.
\newblock \href {https://doi.org/10.18653/v1/2023.acl-long.307} {Evaluating {Open}-{Domain} {Question} {Answering} in the {Era} of {Large} {Language} {Models}}.
\newblock In \emph{Proceedings of the 61st {Annual} {Meeting} of the {Association} for {Computational} {Linguistics} ({Volume} 1: {Long} {Papers})}, pages 5591--5606, Toronto, Canada. Association for Computational Linguistics.

\bibitem[{Kingma and Ba(2017)}]{kingma_adam_2017}
Diederik~P. Kingma and Jimmy Ba. 2017.
\newblock \href {http://arxiv.org/abs/1412.6980} {Adam: {A} {Method} for {Stochastic} {Optimization}}.
\newblock ArXiv:1412.6980 [cs].

\bibitem[{Lindson et~al.(2019)Lindson, Thompson, Ferrey, Lambert, and Aveyard}]{lindson_motivational_2019}
Nicola Lindson, Tom~P Thompson, Anne Ferrey, Jeffrey~D Lambert, and Paul Aveyard. 2019.
\newblock \href {https://doi.org/10.1002/14651858.CD006936.pub4} {Motivational interviewing for smoking cessation}.
\newblock \emph{The Cochrane Database of Systematic Reviews}, 2019(7):CD006936.

\bibitem[{Liu et~al.(2022)Liu, Tao, Liang, Shen, Feng, Huang, and Zhao}]{liu_rethinking_2022}
Chang Liu, Chongyang Tao, Jianxin Liang, Tao Shen, Jiazhan Feng, Quzhe Huang, and Dongyan Zhao. 2022.
\newblock \href {https://aclanthology.org/2022.emnlp-main.729} {Rethinking {Task}-{Specific} {Knowledge} {Distillation}: {Contextualized} {Corpus} as {Better} {Textbook}}.
\newblock In \emph{Proceedings of the 2022 {Conference} on {Empirical} {Methods} in {Natural} {Language} {Processing}}, pages 10652--10658, Abu Dhabi, United Arab Emirates. Association for Computational Linguistics.

\bibitem[{McHugh(2012)}]{mchugh_interrater_2012}
Mary~L. McHugh. 2012.
\newblock \href {https://www.ncbi.nlm.nih.gov/pmc/articles/PMC3900052/} {Interrater reliability: the kappa statistic}.
\newblock \emph{Biochemia Medica}, 22(3):276--282.

\bibitem[{Miller and Rollnick(2012)}]{miller_motivational_2012}
William~R. Miller and Stephen Rollnick. 2012.
\newblock \emph{Motivational {Interviewing}: {Helping} {People} {Change}}.
\newblock Guilford Press.

\bibitem[{Nyamathi et~al.(2010)Nyamathi, Shoptaw, Cohen, Greengold, Nyamathi, Marfisee, de~Castro, Khalilifard, George, and Leake}]{nyamathi_effect_2010}
Adeline Nyamathi, Steven Shoptaw, Allan Cohen, Barbara Greengold, Kamala Nyamathi, Mary Marfisee, Viviane de~Castro, Farinaz Khalilifard, Daniel George, and Barbara Leake. 2010.
\newblock \href {https://doi.org/10.1016/j.drugalcdep.2009.08.021} {Effect of {Motivational} {Interviewing} on {Reduction} of {Alcohol} {Use}}.
\newblock \emph{Drug and Alcohol Dependence}, 107(1):23--30.

\bibitem[{OpenAI(2023)}]{openai_gpt-4_2023}
OpenAI. 2023.
\newblock \href {http://arxiv.org/abs/2303.08774} {{GPT}-4 {Technical} {Report}}.
\newblock ArXiv:2303.08774 [cs].

\bibitem[{Paszke et~al.(2019)Paszke, Gross, Massa, Lerer, Bradbury, Chanan, Killeen, Lin, Gimelshein, Antiga, Desmaison, Kopf, Yang, DeVito, Raison, Tejani, Chilamkurthy, Steiner, Fang, Bai, and Chintala}]{paszke_pytorch_2019}
Adam Paszke, Sam Gross, Francisco Massa, Adam Lerer, James Bradbury, Gregory Chanan, Trevor Killeen, Zeming Lin, Natalia Gimelshein, Luca Antiga, Alban Desmaison, Andreas Kopf, Edward Yang, Zachary DeVito, Martin Raison, Alykhan Tejani, Sasank Chilamkurthy, Benoit Steiner, Lu~Fang, Junjie Bai, and Soumith Chintala. 2019.
\newblock \href {https://papers.nips.cc/paper_files/paper/2019/hash/bdbca288fee7f92f2bfa9f7012727740-Abstract.html} {{PyTorch}: {An} {Imperative} {Style}, {High}-{Performance} {Deep} {Learning} {Library}}.
\newblock In \emph{Advances in {Neural} {Information} {Processing} {Systems}}, volume~32. Curran Associates, Inc.

\bibitem[{Radford et~al.(2019)Radford, Wu, Child, Luan, Amodei, and Sutskever}]{radford_language_2019}
Alec Radford, Jeff Wu, Rewon Child, D.~Luan, Dario Amodei, and Ilya Sutskever. 2019.
\newblock \href {https://www.semanticscholar.org/paper/Language-Models-are-Unsupervised-Multitask-Learners-Radford-Wu/9405cc0d6169988371b2755e573cc28650d14dfe} {Language {Models} are {Unsupervised} {Multitask} {Learners}}.

\bibitem[{Rajbhandari et~al.(2020)Rajbhandari, Rasley, Ruwase, and He}]{rajbhandari_zero_2020}
Samyam Rajbhandari, Jeff Rasley, Olatunji Ruwase, and Yuxiong He. 2020.
\newblock \href {http://arxiv.org/abs/1910.02054} {{ZeRO}: {Memory} {Optimizations} {Toward} {Training} {Trillion} {Parameter} {Models}}.
\newblock ArXiv:1910.02054 [cs, stat].

\bibitem[{Sanh et~al.(2020)Sanh, Debut, Chaumond, and Wolf}]{sanh_distilbert_2020}
Victor Sanh, Lysandre Debut, Julien Chaumond, and Thomas Wolf. 2020.
\newblock \href {http://arxiv.org/abs/1910.01108} {{DistilBERT}, a distilled version of {BERT}: smaller, faster, cheaper and lighter}.
\newblock ArXiv:1910.01108 [cs].

\bibitem[{Shen et~al.(2022)Shen, Perez-Rosas, Welch, Poria, and Mihalcea}]{shen_knowledge_2022}
Siqi Shen, Veronica Perez-Rosas, Charles Welch, Soujanya Poria, and Rada Mihalcea. 2022.
\newblock \href {https://doi.org/10.18653/v1/2022.acl-long.221} {Knowledge {Enhanced} {Reflection} {Generation} for {Counseling} {Dialogues}}.
\newblock In \emph{Proceedings of the 60th {Annual} {Meeting} of the {Association} for {Computational} {Linguistics} ({Volume} 1: {Long} {Papers})}, pages 3096--3107, Dublin, Ireland. Association for Computational Linguistics.

\bibitem[{Shen et~al.(2020)Shen, Welch, Mihalcea, and Pérez-Rosas}]{shen_counseling-style_2020}
Siqi Shen, Charles Welch, Rada Mihalcea, and Verónica Pérez-Rosas. 2020.
\newblock \href {https://aclanthology.org/2020.sigdial-1.2} {Counseling-{Style} {Reflection} {Generation} {Using} {Generative} {Pretrained} {Transformers} with {Augmented} {Context}}.
\newblock In \emph{Proceedings of the 21th {Annual} {Meeting} of the {Special} {Interest} {Group} on {Discourse} and {Dialogue}}, pages 10--20, 1st virtual meeting. Association for Computational Linguistics.

\bibitem[{Tang et~al.(2019)Tang, Lu, Liu, Mou, Vechtomova, and Lin}]{tang_distilling_2019}
Raphael Tang, Yao Lu, Linqing Liu, Lili Mou, Olga Vechtomova, and Jimmy Lin. 2019.
\newblock \href {http://arxiv.org/abs/1903.12136} {Distilling {Task}-{Specific} {Knowledge} from {BERT} into {Simple} {Neural} {Networks}}.
\newblock ArXiv:1903.12136 [cs].

\bibitem[{Taori et~al.(2023)Taori, Gulrajani, Zhang, Dubois, Li, Guestrin, Liang, and Hashimoto}]{alpaca}
Rohan Taori, Ishaan Gulrajani, Tianyi Zhang, Yann Dubois, Xuechen Li, Carlos Guestrin, Percy Liang, and Tatsunori~B. Hashimoto. 2023.
\newblock Stanford alpaca: An instruction-following llama model.
\newblock \url{https://github.com/tatsu-lab/stanford_alpaca}.

\bibitem[{Touvron et~al.(2023)Touvron, Lavril, Izacard, Martinet, Lachaux, Lacroix, Rozière, Goyal, Hambro, Azhar, Rodriguez, Joulin, Grave, and Lample}]{touvron_llama_2023}
Hugo Touvron, Thibaut Lavril, Gautier Izacard, Xavier Martinet, Marie-Anne Lachaux, Timothée Lacroix, Baptiste Rozière, Naman Goyal, Eric Hambro, Faisal Azhar, Aurelien Rodriguez, Armand Joulin, Edouard Grave, and Guillaume Lample. 2023.
\newblock \href {http://arxiv.org/abs/2302.13971} {{LLaMA}: {Open} and {Efficient} {Foundation} {Language} {Models}}.
\newblock ArXiv:2302.13971 [cs].

\bibitem[{Wang et~al.(2019)Wang, Singh, Michael, Hill, Levy, and Bowman}]{wang_glue_2019}
Alex Wang, Amanpreet Singh, Julian Michael, Felix Hill, Omer Levy, and Samuel~R. Bowman. 2019.
\newblock \href {http://arxiv.org/abs/1804.07461} {{GLUE}: {A} {Multi}-{Task} {Benchmark} and {Analysis} {Platform} for {Natural} {Language} {Understanding}}.
\newblock ArXiv:1804.07461 [cs].

\bibitem[{Wang et~al.(2023)Wang, Kordi, Mishra, Liu, Smith, Khashabi, and Hajishirzi}]{wang_self-instruct_2023}
Yizhong Wang, Yeganeh Kordi, Swaroop Mishra, Alisa Liu, Noah~A. Smith, Daniel Khashabi, and Hannaneh Hajishirzi. 2023.
\newblock \href {http://arxiv.org/abs/2212.10560} {Self-{Instruct}: {Aligning} {Language} {Models} with {Self}-{Generated} {Instructions}}.
\newblock ArXiv:2212.10560 [cs].

\bibitem[{Wolf et~al.(2020)Wolf, Debut, Sanh, Chaumond, Delangue, Moi, Cistac, Rault, Louf, Funtowicz, Davison, Shleifer, von Platen, Ma, Jernite, Plu, Xu, Scao, Gugger, Drame, Lhoest, and Rush}]{wolf_huggingfaces_2020}
Thomas Wolf, Lysandre Debut, Victor Sanh, Julien Chaumond, Clement Delangue, Anthony Moi, Pierric Cistac, Tim Rault, Rémi Louf, Morgan Funtowicz, Joe Davison, Sam Shleifer, Patrick von Platen, Clara Ma, Yacine Jernite, Julien Plu, Canwen Xu, Teven~Le Scao, Sylvain Gugger, Mariama Drame, Quentin Lhoest, and Alexander~M. Rush. 2020.
\newblock \href {http://arxiv.org/abs/1910.03771} {{HuggingFace}'s {Transformers}: {State}-of-the-art {Natural} {Language} {Processing}}.
\newblock ArXiv:1910.03771 [cs].

\bibitem[{Wu et~al.(2023)Wu, Balloccu, Reiter, Helaoui, Reforgiato~Recupero, and Riboni}]{wu_are_2023}
Zixiu Wu, Simone Balloccu, Ehud Reiter, Rim Helaoui, Diego Reforgiato~Recupero, and Daniele Riboni. 2023.
\newblock \href {https://aclanthology.org/2023.acl-long.382} {Are {Experts} {Needed}? {On} {Human} {Evaluation} of {Counselling} {Reflection} {Generation}}.
\newblock In \emph{Proceedings of the 61st {Annual} {Meeting} of the {Association} for {Computational} {Linguistics} ({Volume} 1: {Long} {Papers})}, pages 6906--6930, Toronto, Canada. Association for Computational Linguistics.

\end{thebibliography}
\bibliographystyle{acl_natbib}
\appendix
\section{GPT-4 Prompts}
This section shows all of the prompts used with GPT-4 described in this paper.
\label{app:gpt-4prompts}
\begin{table*}[!htb]
\begin{tabular}{|p{2cm}|p{13cm}|} 
 \hline
 Prompt Name & Prompt\\ [0.5ex] 
 \hline
 Simple Reflection Generation & The following is an interaction between a therapist and a client. Act as the therapist and give a reflection to the client's response. The reflection must be a statement and not a question. The reflection must be a rephrasing of the client's response.\\
 \hline
 Complex Reflection Generation & The following is an interaction between you and a user. You are a therapist and the user is someone having smoking issues. Give a SHORT reflection to the user's response. The reflection must be a plausible guess or assumption about the user's underlying emotions, values, or chain of thought. The reflection must be very short. The reflection must be a statement and not a question. Don't always use ``it seems like" or ``it sounds like" or ``you" at the beginning. Don't always use the phrase ``important to you" or ``important for you".\\
 \hline
MI-Adherence & Decide whether the ``reflection" sentence in the following smoking-related conversation meets the standards for Motivational Interviewing. If it does, output ``True"; otherwise, output ``False".\newline
Additionally, a good reflection must:\newline
1. Be a statement, not a question.\newline
2. Not be MI-inconsistent in the following ways: giving advice or information without permission, or confronting the person by disagreeing, arguing, correcting, shaming, blaming, criticizing, labeling, ridiculing, or questioning the person’s honesty, or directing the person by giving orders, commands, or imperatives, or otherwise challenging the person’s autonomy.\newline
3.Not incentivize people to smoke more, or discourage people from quitting smoking.\newline
4.Not exaggerate or understate the sentiment of the sentence to be reflected.\newline
5. Not be factually wrong about smoking.\newline
6. Be grammatically correct.\\

 \hline
 Reflection Type Classification & Decide whether the ``reflection" sentence in the following smoking-related conversation is a SIMPLE or COMPLEX reflection. If it is simple, output ``simple"; otherwise, output ``complex".\newline 
 A simple reflection must be a rephrasing of the client’s response. In contrast, a complex reflection must not be just a rephrasing of the client’s response, but instead a plausible guess or assumption about the user’s underlying emotions, values, or chain of thought.\\
 \hline
\end{tabular}
\caption{\label{tab:gpt-4prompts}
All GPT-4 Prompts
}
\end{table*}

\newpage
\phantom{ }
\newpage
\phantom{ }
\newpage

\section{Fine-tuning Text Format}
This section shows the text formatting this work uses for fine-tuning.
\phantom{text}
\label{app:appendixfinetuningtextformat}
\begin{table}[!htb]
\centering
\begin{tabular}{p{7cm}}
 \hspace{18 mm}\textbf{Simple Reflection Entry} \\
 \hline
 \#\#\# Instruction: \\
 The following is an interaction between a therapist and a client. Act as the therapist and give a reflection to the client's response. The reflection must be a statement and not a question. The reflection must be a rephrasing of the client's response.\\
 \#\#\# Conversation:\\
 Therapist: Now, what is the thing you like least about smoking?\\
 Client: That I have to hide it from my family.\\
 Therapist: You feel the need to keep your smoking habit a secret from your family.\\
 \hline
 \hspace{18 mm}\textbf{Complex Reflection Entry} \\
 \hline
 \#\#\# Instruction: \\
 The following is an interaction between you and a user. You are a therapist and the user is someone having smoking issues. Give a SHORT reflection to the user's response. The reflection must be a plausible guess or assumption about the user's underlying emotions, values, or chain of thought. The reflection must be very short. The reflection must be a statement and not a question. Don't always use "it seems like" or "it sounds like" or "you" at the beginning. Don't always use the phrase "important to you" or "important for you".\\
 \#\#\# Conversation:\\
 Therapist: Now, what is the thing you like least about smoking?\\
 Client: That I have to hide it from my family.\\
 Therapist: You're feeling guilty and secretive about your smoking habit.\\
 \hline
\end{tabular}
\caption{\label{tab:triplet}
Simple and Complex Reflection Dataset Entry Example.
}
\end{table}
\newpage

\section{Hyperparameters}
This section shows the final hyperparameters selected.
\phantom{invisibletext}
\label{app:hyperparameters}
\begin{table}[!htb]
\centering
\begin{tabular}{p{2.5cm}ll}
Model                  & Learning Rate & Batch Size \\
\hline
GPT-2 Small - Simple   & 0.0005        & 32         \\
GPT-2 Medium - Simple  & 0.00005       & 64         \\
GPT-2 Large - Simple   & 0.00005       & 64         \\
GPT-2 XL - Simple      & 0.00005       & 64         \\
GPT-2 Small - Complex  & 0.0005        & 32         \\
GPT-2 Medium - Complex & 0.00005       & 64         \\
GPT-2 Large - Complex  & 0.00005       & 64         \\
GPT-2 XL - Complex     & 0.00005       & 64        
\end{tabular}
\caption{\label{tab:hyperparameters}
Hyperparameters Results for GPT-2 Student Models.
}
\end{table}

\section{GPT-4 Evaluation Precision, Recall, and F1}
\label{app:precisionrecallf1}
This section shows the precision, recall, and F1 scores of the GPT-4 MI-Adherence classifier and the GPT-4 Reflection Type classifier. These scores are calculated by using the human review decisions from Section~\ref{sec:humanreview} as true labels and decisions made by GPT-4 as predicted labels.
\begin{table}[!htb]
\centering
\begin{tabular}{p{3cm}lll}
Model  & Precision & Recall & F1 \\
\hline
GPT-4 MI-A   & 0.967  & 0.935 & 0.951 \\
GPT-4 RT-CLS & 0.835 & 0.789 & 0.811 \\
\end{tabular}
\caption{\label{tab:precisionrecallf1}
Precision, Recall, and F1 scores for GPT-4 Evaluation Models. MI-A and RT-CLS refer to Motivational Inter-
viewing Adherence and Reflection Type Classification
respectively.
}
\end{table}

\end{document}